# ROBOTIC BEHAVIOR PREDICTION USING HIDDEN MARKOV MODELS

**Alan J. Hamlet, Carl D. Crane**
Department of Mechanical and Aerospace Engineering, University of Florida
Gainesville, Florida, USA

## ABSTRACT

*There are many situations in which it would be beneficial for a robot to have predictive abilities similar to those of rational humans. Some of these situations include collaborative robots, robots in adversarial situations, and for dynamic obstacle avoidance. This paper presents an approach to modeling behaviors of dynamic agents in order to empower robots with the ability to predict the agent's actions and identify the behavior the agent is executing in real time. The method of behavior modeling implemented uses hidden Markov models (HMMs) to model the unobservable states of the dynamic agents. The background and theory of the behavior modeling is presented. Experimental results of realistic simulations of a robot predicting the behaviors and actions of a dynamic agent in a static environment are presented.*

## INTRODUCTION

Robots are becoming increasingly autonomous, requiring them to operate in more complicated environments and assess complex phenomena in the world around them. The ability to determine the behavior of humans, other robots, and obstacles efficiently would benefit many fields of robotics. Some such fields include rescue robots, collaborative or assistive robots, multi-robot systems, military robots, and autonomous vehicles. The process of a robot determining the behavior or intent of other agents is inherently probabilistic since the robot has no way of 'knowing' the intent of the other agent. The robot must make a best guess based on its observations of the agent's states. Yet, the same behavior may manifest itself as different sequences in the change of state of the agent being observed. Additionally, robots observe the state of the agent via noisy sensors. This results in a doubly stochastic process where an agent's behavior is probabilistically dependent on the sequence of state changes it undergoes, and the observations seen by the robot depend probabilistically on the state the agent is in. A good model for the behavior of an agent being observed, therefore, would incorporate both of these types of uncertainty. One such model is a hidden Markov model (HMM). A HMM is a statistical model that is a Markov process with states that cannot be directly observed, i.e. they are 'hidden'. The basic theory of hidden Markov models was developed and published in the early 1970's by Baum and his colleagues [1]-[2]. In the following several years HMMs started being applied to speech processing by Jelinek and others [3],[8],[9]. Since the theories were originally published in mathematics journals, HMM techniques were mostly unknown to engineers until Rabiner [13] published his popular tutorial in 1989 thoroughly describing HMMs and some applications in speech processing. Research in other applications has begun to emerge.

In some previous works, there has been a focus on predicting the behaviors of human drivers. In [4] and [5], the authors researched in-vehicle systems that predict the operators' driving intent, but these systems have direct access to the driver's inputs to the vehicle (i.e. steering angle, accelerator depression). In [6] Han and Veloso use hidden Markov models to predict the behaviors of soccer playing robots and utilize the probability of being in certain 'accept' or 'reject' states as the condition for establishing occurrence of a behavior. In another application to robotic soccer, [11] uses 'sequential pattern mining' to look for mappings between currently occurring patterns and stored predicate patterns. In [7] the authors model human trajectories based on position observations at certain time intervals. A HMM approach is used in [10], where the authors study a mobile robot observing interactions between two humans using a camera.

In this paper, a behavior modeling approach using hidden Markov models is introduced that utilizes event based observations. A novel method of determining whether any of a set of non-exclusive behaviors are occurring is described. The performance of these models is then tested in realistic simulations involving two mobile robots in an indoor environment. The observations for the models are made by the robot using a common laser range finder. The remainder of this paper is laid out as follows: first, the background and theory of the HMM behavior models is explained. Next, the method of behavior prediction is introduced, followed by the experimental set-up and simulation description. Finally, the results of the simulations are presented and discussed.

## BEHAVIOR MODELING

In this paper, 'behaviors' are defined as a temporal series of state changes of a dynamic agent. The states of the dynamic agent are assumed to be not directly observable, but having observations that are correlated to the agent's state in a probabilistic manner. This is a reasonable assumption because when the higher level intent or goal of a dynamic agent is to be inferred, the relevant states would be the agent's internal or mental states. These states are not directly observable but will result in measurable physical phenomena that are correlated to the agent's state. The behavior dictates how the agent changes from one state to another. When behaviors are defined in this way, hidden Markov models are ideal representations for them.

There are several elements used to describe a HMM. The number of states, $N$, is the number of distinct, finite states in the model. Individual states are represented as

$$S = \{S_1, S_2, \ldots, S_N\}$$

The current state is $q_t$. Although the observations in hidden Markov models can be continuous random variables, in this paper only behavior models that have a finite number of discrete observation symbols are considered. Therefore, $M$ is the number of possible observation tokens in the model where the individual observation tokens are denoted as

$$V = \{v_1, v_2, \ldots, v_M\}.$$

The state transition matrix

$$A = \{a_{ij}\}$$

is the $N \times N$ matrix of probabilities of switching from one state to another, i.e.

$$a_{ij} = P(q_{t+1} = S_j | q_t = S_i).$$

It should be noted that HMMs assume the system follows the Markov, or memoryless property. The assumption of the Markov property is that the probability of the next state of the system depends only on the current state of the system and is independent of the previous system states, i.e.

$$P(q_t = S_j | S_1, S_2 \ldots S_{t-1}) = P(q_t = S_j | S_{t-1})$$

The observation symbol probability distribution for state j is

$$b_j(k) = P(v_k \text{ at time } t \mid q_t = S_j)$$
$$\text{for } 1 \leq j \leq N \text{ and } 1 \leq k \leq M$$

In other words, the probability of seeing observation symbol $v_k$ at time $t$, given the state at time $t$ is $S_j$. In the discrete model, this yields an $N \times M$ matrix of observation probabilities.

The final element of HMMs is the initial state distribution,

$$\pi_i = P(q_1 = S_i) \text{ for } 1 \leq i \leq N.$$

The entire model can be specified by the three probability measures $A$, $B$, and $\pi$ so the concise notation for the model

$$\lambda = (A, B, \pi)$$

is used. The goal is to determine if a series of observations

$$O = O_1 O_2 \ldots O_T$$

was generated by a behavior modeled as $\lambda = (A,B,\pi)$, where each $O_t$ is an observation symbol in $V$. Throughout this paper the 'agent' is the dynamic object that is being observed and whose behavior is to be determined. The agent could be a human, robot, or moving obstacle. The 'robot' is the observer trying to determine the agent's intent.

In typical applications, observation symbols for HMMs are measured at regular time intervals. To take advantage of the discrete nature of the models used here and reduce the length of the observation sequences (and therefore the required computation), observations symbol generation is event-based. Only when an event triggers the value of an observation to change is a symbol generated. The subscript $t$ is used to denote the sequential time step, but the temporal spacing between observations can be highly irregular.

## BEHAVIOR PREDICTION

Using the form of the behavior model described above, models can be trained to match observations generated from a behavior of interest. While there is no analytical solution for optimizing the model parameters, $\lambda = (A,B,\pi)$, to maximize $P(O|\lambda)$, iterative techniques can be used to find a local maxima. A modified version of Expectation-Maximization known as the Baum Welch algorithm is a commonly used method of doing this and is the method used in this study. Since the solution is only a local maximum, care must be taken in selecting initial conditions. Once the model is trained on a series of observation sequences, the model can be used to calculate the likelihood of a new sequence being generated by the model. This is where the strength of HMMs lie, since by using dynamic programming techniques this can be calculated in just $O(N^2T)$ calculations. To do so, a variable known as the forward variable is defined:

$$\alpha_t(i) = P(O_1 O_2 \ldots O_t, q_t = S_i | \lambda)$$

Where $\alpha_t(i)$ is the joint probability of the observation sequence up to time $t$ and the state being $S_i$ at time $t$, given the model, λ. The probability of the entire sequence given the model, $P(O|\lambda)$, can be found inductively in three steps as follows

1) $\alpha_1(i) = \pi_i b_i(O_1), \quad 1 \leq i \leq N$

2) $\alpha_{t+1}(j) = \left[\sum_{i=1}^{N} \alpha_t(i) a_{ij}\right] b_j(O_{t+1}),$
   $1 \leq t \leq T-1 \quad 1 \leq j \leq N$

3) $P(O|\lambda) = \sum_{i=1}^{N} \alpha_T(i)$

Here, the forward variables are first initialized as the joint probabilities of the state $S_i$ and the observation $O_1$. To calculate the next forward variable, the previous forward variables are multiplied by the transition probabilities and summed up (to find the probability of transitioning to state $S_j$) before being multiplied by the next observation probability. Finally, by summing up the forward variables over all the states at the final time step, the state variable is marginalized out and the probability of the entire observation sequence is obtained. In the same manner, by summing up the forward variables at the current time step, the probability of the current partial observation sequence can be found.

If the goal is to discriminate between a finite set of $Z$ known independent behaviors, modeled as $\lambda_i$, $1 \leq i \leq Z$, then the probability can be normalized as

$$P(O|\lambda_i) = \frac{\sum_{i=1}^{N} \alpha_T(i)}{\sum_{j=1}^{Z} \sum_{i=1}^{N} \alpha_T^j(i)}$$

so that $\sum_{i=1}^{Z} P(O|\lambda_i) = 1$ and the most likely behavior is simply

$$\operatorname*{argmax}_{1 \leq i \leq Z} P(O|\lambda_i).$$

This same process can also be done online on partial observation sequences so that the most likely occurring behavior can be updated after every time step.

While the above procedure is optimal when models for all possibly occurring behaviors are available (and are independent of one another), this is rarely the case. Also, it would be beneficial to be able to add or remove behaviors without modifying software and to be able to determine if none, or multiple, of the modeled behaviors are occurring. Therefore, a novel method of obtaining the likelihood of behaviors is introduced.

In itself, the probability of a partial observation sequence for a given model does not convey very much information. This is because, depending on the number of possible observation sequences, even the most likely sequence of observations can have a very low probability. Therefore, in order to determine the likelihood of a behavior model given a partial observation sequence, the probability of the current observation sequence is normalized with respect to the probability of the most likely observation sequence of equal length. Then, that likelihood is scaled by the predicted fraction of the behavior that has been observed.

$$L(\lambda_i | O_1 O_2 \ldots O_t) = \frac{\sum_{i=1}^{N} \alpha_T(i)}{\max_{O_{1:t}} P(O_{1:t}|\lambda_i)} \frac{t}{T_i}$$

Behaviors are not assumed to be of equal length, therefore, the length of an observation sequence corresponding to behavior $i$ is denoted as $T_i$. For each behavior, the normalizers were pre-computed as $\max_{O_{1:t}} P(O_{1:t}|\lambda_i)$, $for\ 1 \leq t \leq T_i$.

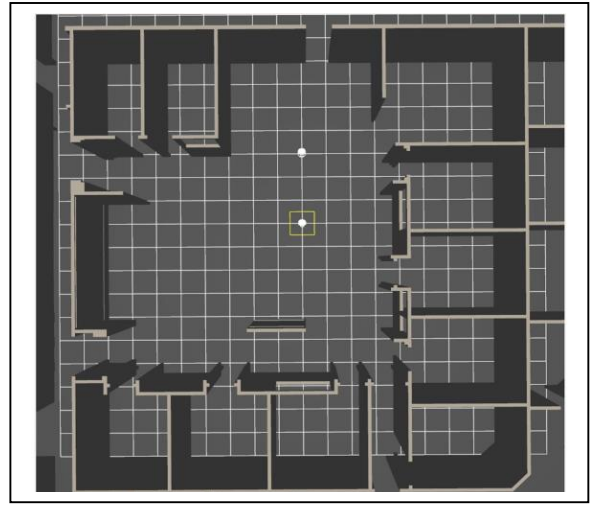

Figure 1. Aerial view of the simulated office environment.

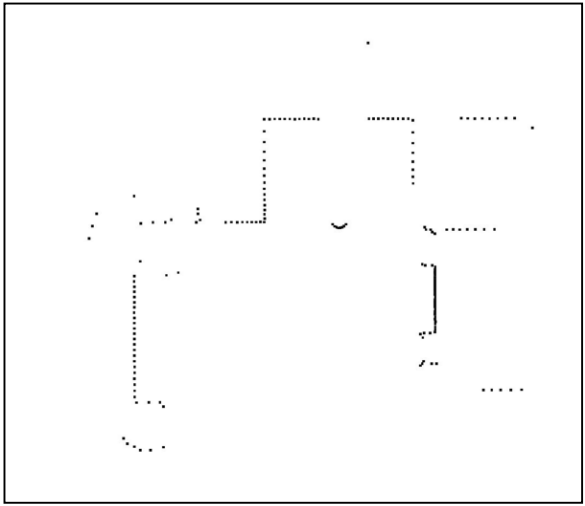

Figure 2. Visualization of a laser scan of the environment containing a dynamic agent.

## EXPERIMENTAL SET UP

The behavior models described above were tested in realistic simulations using the open source software ROS (Robot Operating System) and Gazebo, an open source multi-robot simulator. The simulations were conducted using two Turtlebots in a simulated office environment, shown in Figure 1 with the observing robot toward the center of the figure and the robot representing the dynamic agent toward the top of the figure. The following sections describe the simulation environment and the software implemented in order to test the behavior models presented in the previous section.

### Simulation Environment

The software used for the experimental simulations, Gazebo, accurately simulates rigid-body physics in a 3D environment. It also generates realistic sensor feedback and interactions between three-dimensional objects. An aerial view of the office environment used for the simulations is shown in Figure 1 above. The observations for the behavior models were generated using the robot's planar laser range finder. A visualization of the laser range finder's output is shown in Figure 2. The robot is facing the direction corresponding to up in Figures 1 and 2, and the dynamic agent as seen by the robot can be seen in Figure 2. During the simulations, the sensor data generated by the simulator was sent to ROS for processing and running the behavior recognition algorithms, all in real-time. Both Turtlebots were equipped with an inertial measurement unit and encoders for measuring wheel odometry. All sensor data (laser, IMU, and wheel encoders) had added Gaussian noise to mimic real-world uncertainty.

### ROS Software

While some software for operating Turtlebots is available open source, all the software for the behavior recognition algorithms was developed, along with the necessary peripheral programs, in C++. The observing robot tracked the dynamic agent by looking for the known robot geometry in each laser range scan. The partial view of the circular robot in the office environment can be seen in the visualization of the laser range scan in Figure 2. The maximum range of the laser range finder was 10 meters and the tracker could identify the robot at a range of up to 7.5 meters. If the robot was identified in a scan, the robot's absolute x-y coordinates were calculated and used as input to a Kalman filter which estimated the velocity. Using the position and velocity information as input to the behavior models, the likelihood of each models occurrence was calculated. The dynamic agent executed behaviors using a PID controller using odometry data as feedback. The odometry data was obtained via an Extended Kalman Filter that fused wheel encoder data with IMU data. A program for generating observations read in the position and velocity of the dynamic agent and published observations when events triggering them occurred. Then behavior recognizers for each of the modeled behaviors determined the likelihood of their respective behavior and output the data in real time.

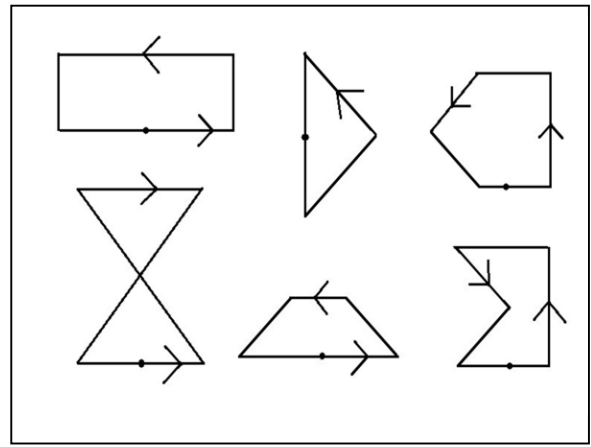

Figure 3. Example trajectories from the six tested behaviors.

## RESULTS

To test the behavior models, six behaviors based on polygonal trajectories were used. The observations used in the models for these trajectories consisted of the change in velocity direction of the dynamic agent. Examples of the trajectories are shown in Figure 3. Going clockwise from top-left in Figure 3, the behaviors are referred to as rectangle, triangle, convex box, concave box, trapezoid, and hourglass.

The tests performed consisted of having the dynamic agent execute each of the six behaviors 10 times while behavior recognizers for all six of the models were running. Each one of the behaviors was randomized in three ways. First, the dynamic agent executed the trajectories in different orientations every time by randomly selecting an initial direction. Second, the side lengths of the polygonal trajectories were varied by randomly selecting a scaling factor between 0.5 and 1.5. Finally, the trajectories were executed in both clockwise and counter-clockwise directions. In addition, the dynamic agent executed the behaviors based on a PID controller receiving feedback from noisy sensors, increasing the variance in trajectories for the same behavior.

The average outputs of the behavior recognizers for each of the six behaviors are graphed in Figures 4-9 as functions of the percent of the behavior executed. The percent of a behavior executed was determined by average distance traveled at the time of the output divided by the total trajectory length. For clarity, only the data for behavior models having non-zero likelihood are shown. In half the cases, the true behavior has the highest likelihood starting from the first observation. In every case, the true behavior has the highest likelihood by the time 40% of the behavior has been executed.

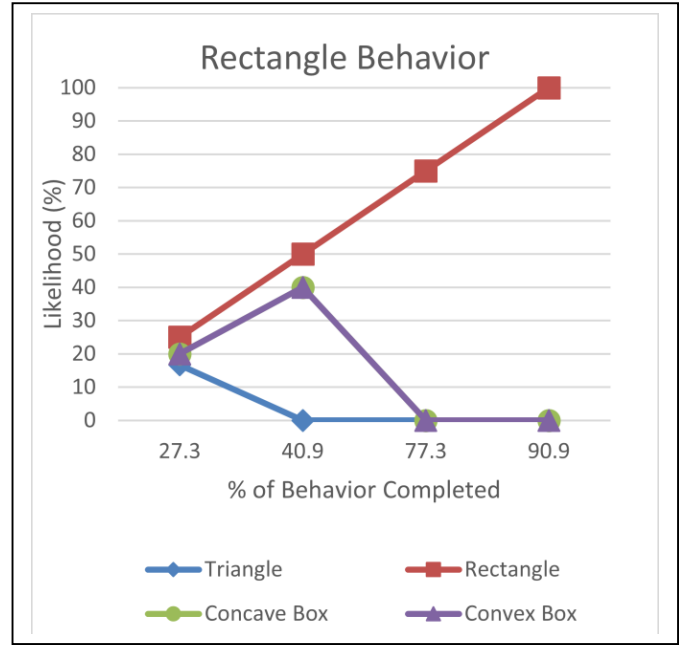

Figure 5. Experimental results for the rectangle behavior.

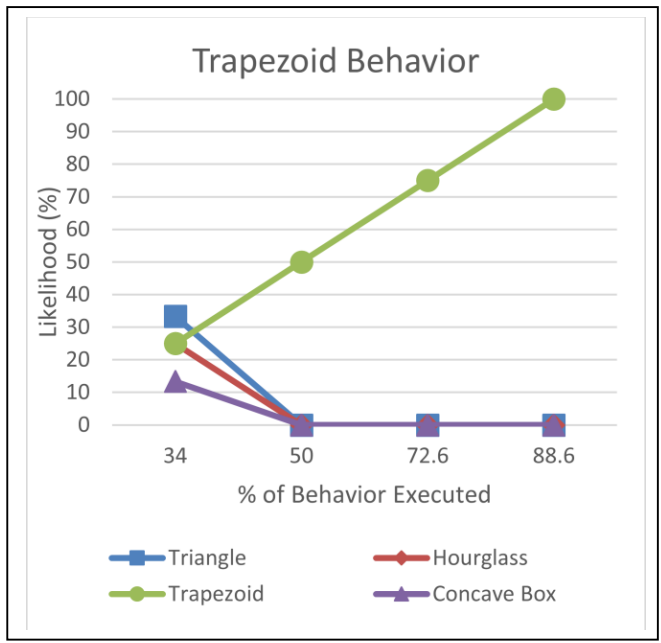

Figure 4. Experimental results for the trapezoid behavior.

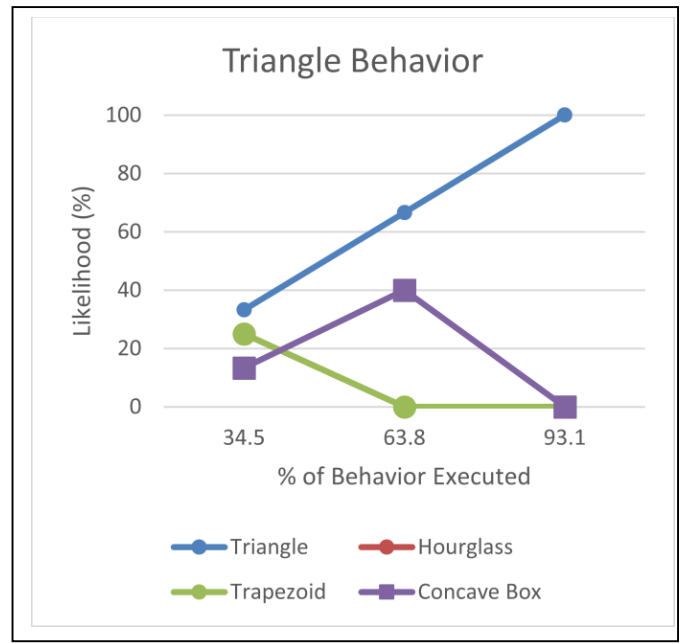

Figure 6. Experimental results for the triangle behavior.

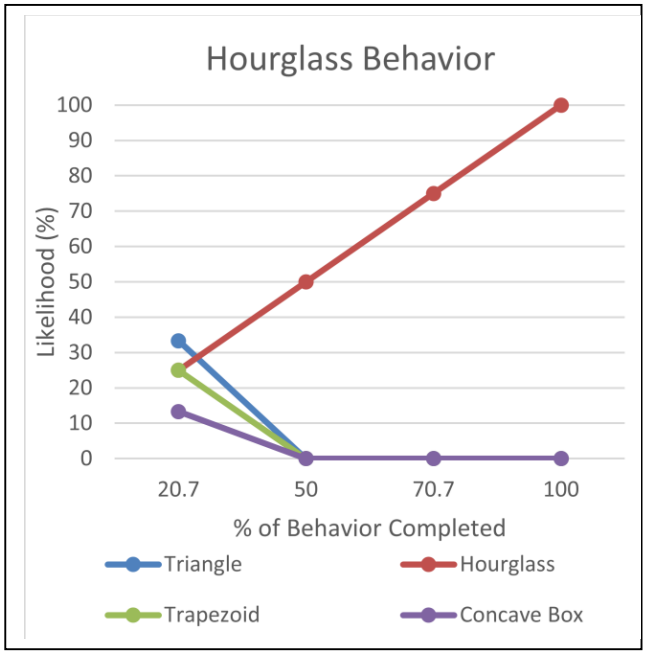

Figure 7. Experimental results for the hourglass behavior.

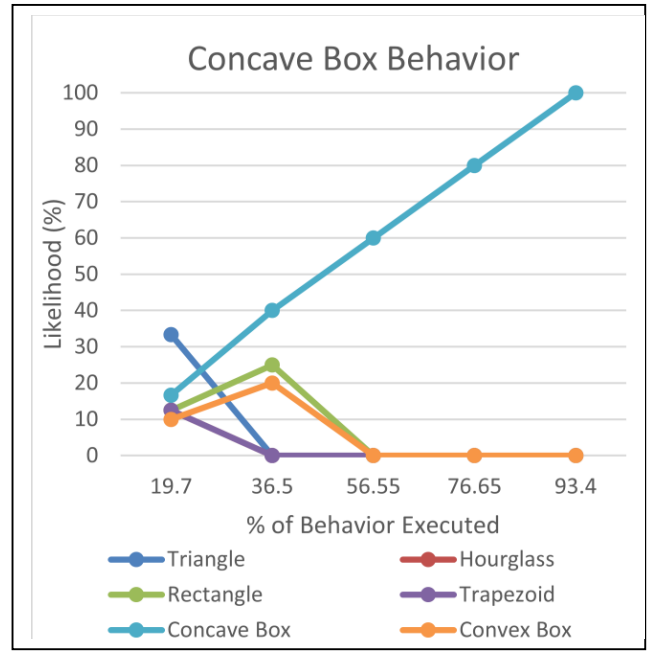

Figure 9. Experimental results for the concave box behavior.

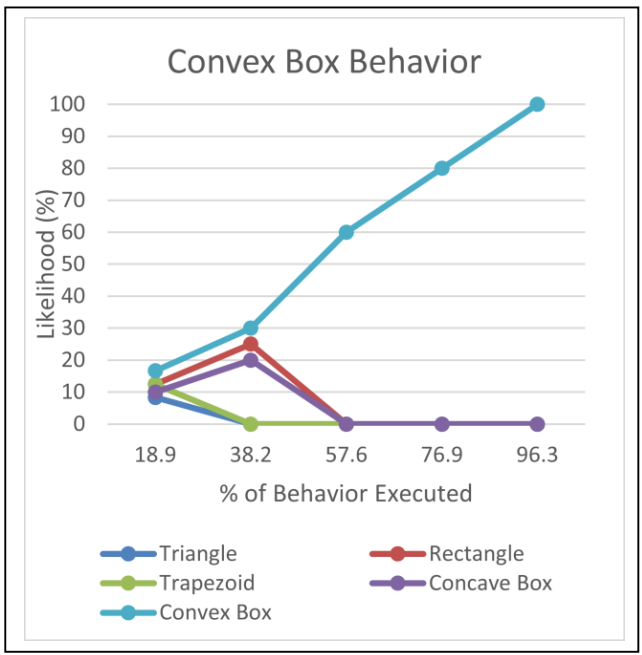

Figure 8. Experimental results for the convex box behavior.

**CONCLUSIONS AND FUTURE WORK**

This paper presented a method of autonomous behavior prediction based on hidden Markov models. A procedure for performing the online classification problem of selecting the most likely behavior from a set of mutually exclusive behaviors was introduced. Next, the procedure was extended in order to determine the individual likelihood of behaviors for the problem of having an incomplete set of potentially dependent behaviors. The method uses event-based observation models and measures the similarity between the current observations and the 'ideal' observations for a given behavior. The models were tested in simulation using two robots in a static environment. The results showed very good prediction of the simple behaviors tested. All models output the highest likelihood for the true behavior by the time 40% of the behavior was executed, and in five out of 6 behaviors, before 25% of the behavior was executed.

In future work, more complex behaviors with realistic applications will be modeled. The models will be expanded to include continuous observation densities, and the models will be tested on physical hardware as well as in simulation.